\documentclass{article} % For LaTeX2e
\usepackage{iclr2025_conference,times}
\usepackage{amsmath,amsfonts,bm}

\def\eqref#1{equation~\ref{#1}}
\def\1{\bm{1}}

\DeclareMathAlphabet{\mathsfit}{\encodingdefault}{\sfdefault}{m}{sl}
\SetMathAlphabet{\mathsfit}{bold}{\encodingdefault}{\sfdefault}{bx}{n}

\usepackage{algorithm}
\usepackage{hyperref}
\usepackage{url}
\usepackage{enumitem}
\usepackage{graphicx} 
\usepackage{algpseudocode}
\usepackage{tabularx}
\usepackage{booktabs}
\usepackage{multirow}
\usepackage{acronym}

\acrodef{PPO}{proximal policy optimization}
\acrodef{DPO}{direct preference optimization}
\acrodef{SFT}{supervised fine-tuning}
\acrodef{DAP}{direct alignment from preference}
\acrodef{SiLC}{sequence likelihood calibration}
\acrodef{IPO}{identity policy optimisation}
\acrodef{AI}{Artificial Intelligence}
\acrodef{LLM}{Large Language Model}

\newcommand{\passage}[1]{%
    \noindent\textbf{#1} % 使用粗体显示文本
}

\usepackage{color, xcolor}
\usepackage{subcaption}
\usepackage{pifont}
\usepackage{textcomp}
\usepackage{hhline}
\usepackage{array} % 用于高级表格布局
\usepackage{colortbl} % 用于表格颜色
\usepackage{makecell}
\usepackage[utf8]{inputenc}
\usepackage{wrapfig}
\title{Beyond Scalar Reward Model: Learning Generative Judge from Preference Data} 
\author{Ziyi Ye\textsuperscript{1}, Xiangsheng Li\textsuperscript{2}, Qiuchi Li\textsuperscript{3}, Qingyao Ai\textsuperscript{1}, Yujia Zhou\textsuperscript{1}, Wei Shen\textsuperscript{2}, \\ 
	\textbf{Dong Yan}\textsuperscript{2}, \textbf{Yiqun Liu}\textsuperscript{1}
	\\ 
	\textsuperscript{1}Department of Computer Science and Technology, Tsinghua University \\ \textsuperscript{2}Baichuan AI \;
	\textsuperscript{3}University of Copenhagen \\
	ye-zy20@mails.tsinghua.edu.cn, \; lixsh6@gmail.com, \; qiuchi.li@di.ku.dk, \;aiqy@tsinghua.edu.cn, \\
	zhouyujia@mail.tsinghua.edu.cn, \;weyshioncn@gmail.com, \;sproblvem@gmail.com, \\yiqunliu@tsinghua.edu.cn
}

\iclrfinalcopy % Uncomment for camera-ready version, but NOT for submission.

\begin{document}

\maketitle
\pagestyle{fancy}
\fancyhf{}
\fancyhead{} % 清除页眉

\begin{abstract}

Learning from preference feedback is a common practice for aligning large language models~(LLMs) with human value.
Conventionally, preference data is learned and encoded into a scalar reward model that connects a value head with an LLM to produce a scalar score as preference or reward.
However, scalar models lack interpretability and are known to be susceptible to biases in datasets. 
This paper investigates leveraging the generation capability of LLMs to address both limitations in one shot. 
Specifically, we prompt the pre-trained LLM to generate positive and negative judgments, both supported with rationales in natural language form. 
The self-generated contrastive judgment pairs are used to train the generative judge with Direct Preference Optimization~(DPO). 
This proposal of training the generative \textbf{J}udge using self-generated \textbf{Con}trastive judgments (Con-J) ensures natural interpretability due to the generated rationales together with the judgments, as well as high robustness against bias without the need for an additional reward head. 
Experimental results show that the performance of Con-J is comparable to the scalar reward model trained on the same collection of preference data, and demonstrate its superior interpretability and robustness in encoding human preferences.

\end{abstract}

\section{Introduction}
As \ac{AI} systems advance with the emergence of \acp{LLM}, it is crucial to ensure they align with human instructions, values, and ethics.
\acp{LLM} alignment is generally achieved by learning from preference data that compares pairs of responses to a question~\citep{rafailov2024direct,christiano2017deep,liu2020learning}.
However, collecting high-quality human preference data is both time-consuming and costly. 
In practice, the construction of preference datasets often involves scaling with a combination of human and AI-generated feedback \cite{lee2023rlaif, hou2024chatglm}. 
An additional advantage of AI feedback is its ability to be incorporated in real-time, which facilitates online algorithms such as iterative-DPO \cite{xiong2024iterative, xu2023some} and online-DPO \cite{guo2024direct}. 
Therefore, it is crucial to develop an efficient and accurate AI-based preference model that aligns with human values.

\begin{figure}[t]
\begin{center}
\includegraphics[width=1.0\linewidth]{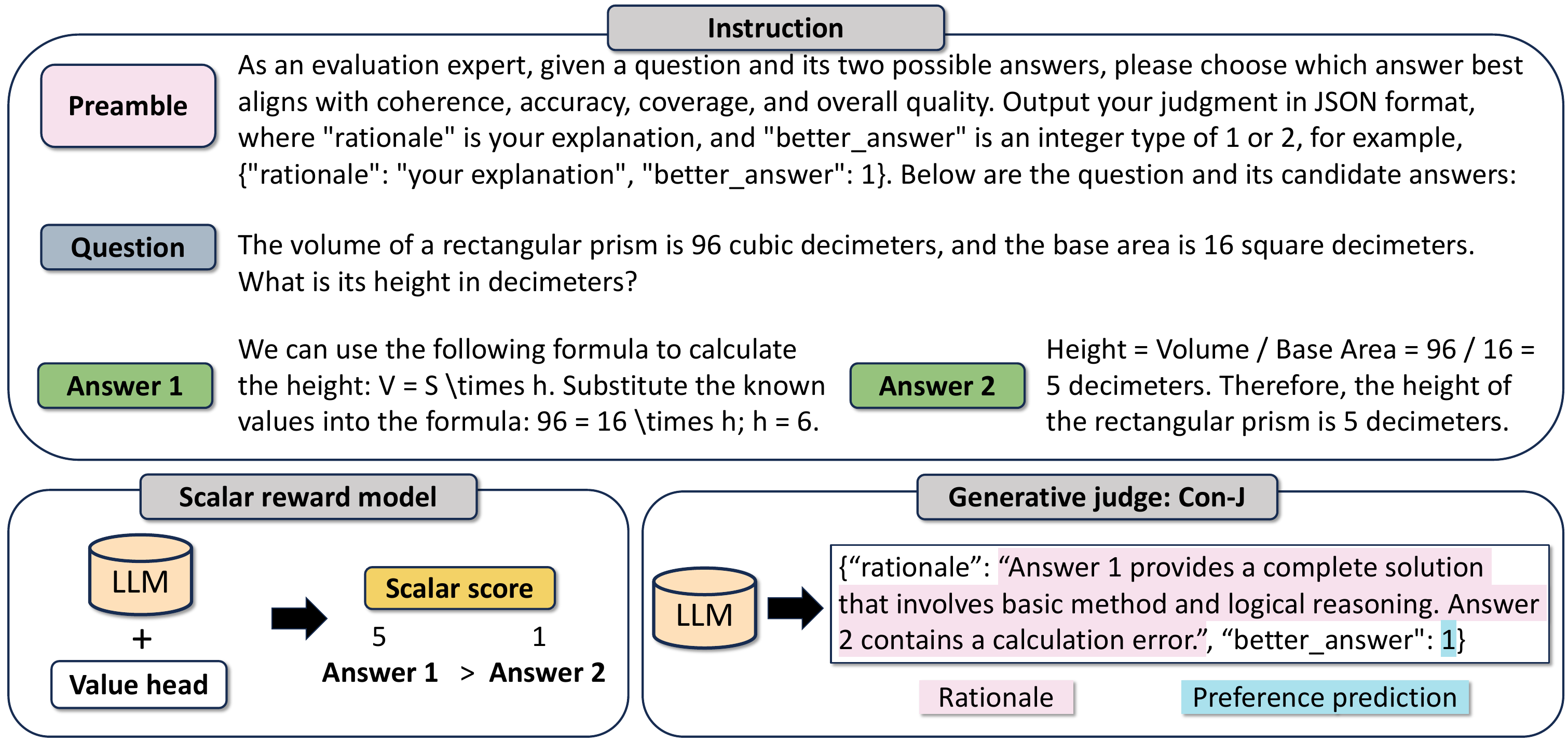}
\end{center}
\caption{Top: Examples of a preamble, a question, a pair of answers, and the corresponding judgment~(see the detailed version in Table~\ref{tab:prompt example}). Bottom: Illustrations of a scalar reward model and the proposed Con-J for preference judgment. \label{fig:introduction}}
\end{figure}

To obtain such preferences, industrial practices have used scalar models \cite{hou2024chatglm} that concatenate the pre-trained LLM with a value head to generate scalar scores for the responses.
However, the scalar model suffers from limitations, particularly in the following aspects: 
(i) \textit{Lack of interpretability}: 
Beyond scalar scores, additional rationales are crucial to enhance the reliability of judgments and facilitate human involvement in the evaluation loop.
(ii) \textit{Susceptibility to bias}: 
It is prone to capturing the biases present in the preference dataset rather than human values. 
For example, when the majority of positive answers in preference datasets are longer sentences, the learned LLM will likely favor more verbose answers~\citep{huang2024empirical}.

To address the above limitations, we propose \textbf{Con-J}, which trains a generative \textbf{J}udge using its self-generated \textbf{Con}trastive judgments~(see Figure~\ref{fig:introduction}).
Con-J leverages the LLM's pre-existing judgment abilities~\cite{zelikman2022star} and bootstraps its capacity to generate more accurate judgments. 
As shown in Figure~\ref{fig:procedure}, Con-J consists of three steps:
$\textbf{(Judgment Sampling)}$~Sample several judgments from a pre-trained LLM by prompting it with a query and a pair of candidate answers. 
$\textbf{(Judgment Filtering)}$~Leverage the true preference annotations to construct contrastive judgment pairs, i.e., judgments with correct or incorrect preference. 
$\textbf{(Training)}$~Train Con-J from the pre-trained LLM based on these contrastive judgments using Direct Preference Optimization~(DPO).
% The above steps enable Con-J to be trained on its self-generated data under the supervision of preference data.
% 可以再加一些ablation，包括不加sft，不加rationale

The design of Con-J differs from existing methods for enhancing the capabilities of generative judges~(or LLM-as-a-judge)~\citep{li2023generative,kim2024prometheus,park2024offsetbias}. 
These methods typically depend on external models~(particularly GPT-4) or algorithmic schemes to produce high-quality instruction-tuning datasets. 
In contrast, Con-J directly learns from preference data using a self-bootstrapping approach similar to that of scalar reward models. 
As LLMs become more powerful, aligning them with high-quality judgments becomes more difficult since humans may not always be able to write superior judgments.
Instead, Con-J offers a new way by eliciting what the LLM already knows, supervised by human preferences, which are much easier to obtain than high-quality judgments.

We train and evaluate Con-J on self-built commercial datasets across three domains: Text Creation, Math, and Code and a series of publicly available datasets and benchmarks.
Our findings indicate that Con-J not only significantly outperforms the scalar model in the Text Creation task and achieves comparable performance in Math and Code but also that its performance, when trained on domain-specific data, will significantly surpass that of GPT-4o.
Additionally, Con-J trained on publicly available datasets achieves comparable performance with GPT-4o and surpasses a series of existing open-source models.
As a generative judge, Con-J can generate rationales to support its preference prediction. 
We evaluate the correctness of these rationales and find that as the accuracy of preference predictions improves, the correctness of the rationales also increases.
Additionally, in a synthetic experiment, we found that Con-J is less susceptible to dataset biases, which we attribute to its generative training target and its feature to generate rationales simultaneously.
To facilitate further research and development within the community, We release the training process and model weights of Con-J trained on publicly available datasets at \href{https://huggingface.co/ZiyiYe/Con-J-Qwen2-7B}{https://huggingface.co/ZiyiYe/Con-J-Qwen2-7B}.

To summarize, our contributions are:
\begin{enumerate}[itemsep=0.0ex,topsep=0.0ex]
    \item We propose Con-J, an approach that trains a generative judge using a self-bootstrapped technology to learn from preference data.
    \item We show that Con-J can offer more accurate rationales during preference learning.
    We also provide theoretical motivation and empirical evidence showing that Con-J can be more robust to dataset biases by training with rationales.
    \item We test the performance of Con-J in commercial datasets and publicly available benchmarks.
    Con-J outperforms the scalar models and a series of existing generative judges.
\end{enumerate}

\section{Preliminary}
\subsection{Task definition}
Given a question or prompt $q$ and a pair of assistant responses $a^1$ and $a^2$, the task is to judge the preference between $a^1$ and $a^2$.
To accomplish this, we train the model using an existing preference dataset \( D = \{(q, a^-, a^+)_i\}_{i=1}^N \), where $a^+$ is a preferred answer compared to $a^-$.
The model's performance is subsequently evaluated on a separate, non-overlapping preference dataset by measuring the accuracy of its preference judgments.

\subsection{Scalar Model}
The most common practice for getting the preference judgment is to use a scalar model~(SM) similar to the reward model in the RLHF stage~(\cite{hou2024chatglm}).
The SM predicts numerical scores $r(q,a)$ for $a\in \{a^1,a^2\}$ and judges the preference by comparing $r(q,a^1)$ and $r(q,a^2)$.
It is typically initialized by concatenating a pre-trained LLM with a randomly initialized shallow MLP head. 
The most widely used training objective for the SM follows the Bradley-Terry model, which maximizes the probability of $a^+$ being preferred:
\begin{equation}
    P(a^+\succ a^-|q)= \frac{\exp(r(q,a^+))}{\exp(r(q,a^+)) + \exp(r(q,a^-))} = \sigma\big(r(q,a^+)- r(q,a^-)\big)
\end{equation}
where $\sigma$ is the sigmoid function.

The above-mentioned SM utilizes the prompt $q$ and a single answer $a$ as input, which we denote as a pointwise SM.
In addition, existing research has investigated a pairwise variant that uses a pair of candidates as input, i.e., $r(q, a^1, a^2)$~\citep{jiang2023llm}. 
The pairwise vanilla reward formalizes the preference probability of $a^+$ as:

\begin{equation}
P(a^+ \succ a^- | q) = \sigma(r(q,a^+,a^-)).
\end{equation}

To train the above-mentioned pointwise and pairwise SM $r$, we maximize the log-likelihood of the preferences by minimizing the following loss function:
\begin{equation}
\ell_R(r) = - \sum_{(x, a^+, a^-)} \log p_r(a^+ \succ a^- \mid x) = \begin{cases} 
    - \sum_{(x, a^+, a^-)} \log \sigma(r(x, a^+) - r(x, a^-)) & \text{(pointwise)} \\
    - \sum_{(x, a^+, a^-)} \log \sigma(r(x, a^+, a^-)) & \text{(pairwise)}
\end{cases}
\end{equation}

\section{Improving Generative Judge by Training on Contrastive Judgments}

\begin{figure}[t]
\begin{center}
%\framebox[4.0in]{$\;$}
\includegraphics[width=1.0\linewidth]{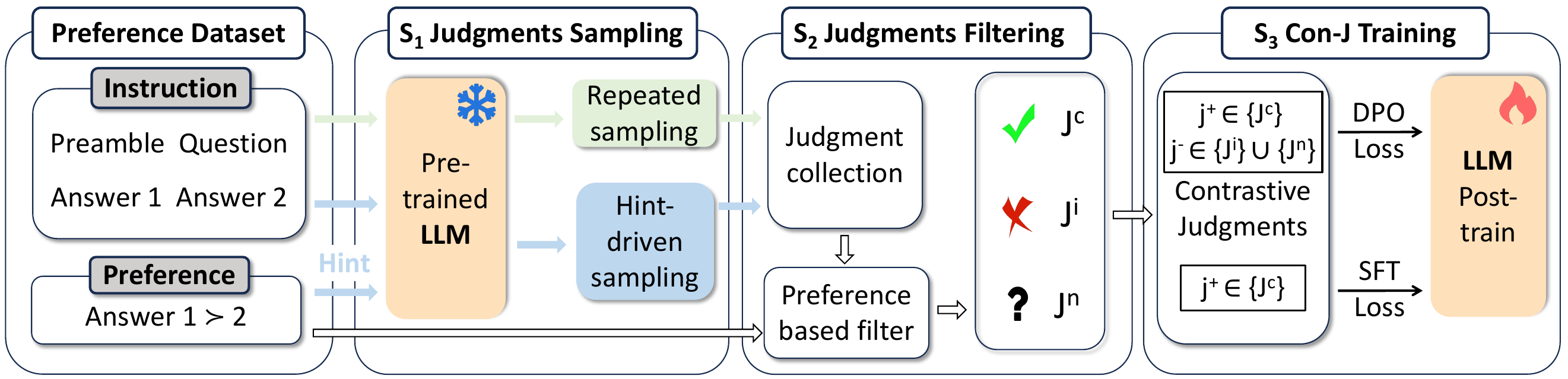}
\end{center}
\caption{The steps for constructing Con-J with a preference dataset that includes preference annotations for a pair of answers to a question.
$S_1$: Prompt Con-J to generate multiple judgments for a pair of answers by repeated sampling and hint-driven sampling. 
$S_2$: Bootstrap contrastive judgment pairs by filtering with true preference labels. A contrastive judgment pair consists of one judgment with the correct preference and the other either with an incorrect preference or does not explicitly indicate a preference.
$S_3$: Train Con-J using the DPO loss on contrastive judgments and the SFT loss on positive judgments.  \label{fig:procedure}}
\end{figure}

Instead of using a scalar model for preference judgment, we propose to leverage the LLM itself to make preference judgments~(\cite{guo2024direct,lee2023rlaif}).
Given the question $q$ and a pair of answers $a^1$ and $a^2$, we construct a prompt $p$ by concatenating a preamble with $q$, $a^1$, and $a^2$.
The preamble is an instruction that describes the task and asks an LLM $\pi$ to act as a judge~(see examples in Fig.~\ref{fig:procedure} and Appendix~\ref{tab:prompt example}). 
Then the LLM generates natural language judgments $j=\pi(p)$, which contains the judgment as well as the rationale in a JSON style: a key named \texttt{"rationale"} includes a step-by-step explanation and verification of the answers, and another key named \texttt{"better\_answer"} indicates the LLM's binary judgment.  

As shown in Figure~\ref{fig:procedure}, the construction of Con-J consists of three steps: \textit{judgment sampling}, \textit{judgment filtering}, and \textit{Con-J training}.

\textit{Judgment sampling:} We construct contrastive judgment pairs by prompting the LLM to generate multiple judgments. 
As shown in Figure~\ref{fig:procedure}, this is achieved by (1)~\textit{repeated sampling} and (2)~\textit{hint-driven sampling}.

Repeated sampling prompts the LLM to generate multiple outputs from the same prompt, each utilizing a different random seed during the generation process. 
However, the LLM may produce only one-sided judgments (i.e., all judgments preferred $a^1$ or $a^2$) across all repeated samples. 
In such cases, we cannot construct contrastive judgment pairs with repeated sampling. 
Therefore, we propose hint-driven sampling to compel the LLM to generate judgments that favor specific answers.
Essentially, the LLM is provided with an explicit indication of which answer is better, and is instructed to generate the judgment accordingly in the same JSON format as above. 
The prompt template for hint-driven sampling is provided in Table~\ref{tab:prompt hint}.
By manipulating the hint, we can get a contrastive judgment pair for any prompt input.

% \subsubsection{Training with Direct Preference Optimization}

\textit{Judgment filtering:} 
We denote the outputs from repeating sampling as $M(p)$.
$M(p)$ can potentially include both ``positive'' and ``negative'' judgments. 
A positive judgment indicates the judgment corresponding to the keyword \texttt{"better\_answer"} is correct~($j^+$), while a negative judgment~($j^-$) indicates the judgment is incorrect~($j^i$) or the model doesn't explicitly indicate its preference~($j^n$). 
Contrastive judgment pairs $\{(j^+, j^-)\}$ are hence constructed as the direct product of the positive judgment set $M(p)^+$ and negative judgment set $M(p)^-$. 
We set the number of repeated samplings to 8, allowing for the construction of up to 4 pairs (in the optimal case, there exist 4 positive and 4 negative judgments among the 8). 
For hint-driven sampling, we prompt the LLM with one correct and one incorrect hint and construct one pair from them. 
The detailed sampling and filtering process is outlined in Algorithm~\ref{alg1}.

\textit{Con-J training:} Based on the constructed judgment pairs  $D^J=\{(q, a^+, a^-, j^+,j^-)_i\}_{i=1}^K$, we train the LLM $\pi$ with a \ac{DPO} loss function:
\begin{equation}
\ell^{\text{DPO}} = - \sum_{(p, j^+, j^-)} \log \sigma\left[\eta \log \frac{\pi({j}^+|p)}{\pi_0({j}^+|p)} - \eta \log \frac{\pi({j}^-|p)}{\pi_0({j}^-|p)}\right]
\end{equation}
where $\pi_0$ is the reference model initialized as the base LLM and remains untrained. 
Following existing practice~\cite{liu2024provably,hong2024orpo,pal2024smaug}, the DPO also fuses a small weight of \ac{SFT} loss to help mitigate the overoptimization issue, which can be formulated as:  
\begin{equation}
    \ell^{\text{SFT}} = - \sum_{(p, j^+)} \log \pi({j}^+|p)
\end{equation}
Then we linearly combine the DPO loss and the SFT loss with a small weight $\alpha$:
\begin{equation}
    \ell^{\text{final}} = \ell^{\text{DPO}} + \alpha * \ell^{\text{SFT}}
\end{equation}

% In practice, we observed a decrease in rewards for both positive and negative judgments. 
% To mitigate this issue, we used a \ac{SFT} loss for positive samples following the existing \ac{DPO} training practice~\cite{liu2024provably,hong2024orpo,pal2024smaug}.

\passage{DPO training promotes distinguishing between answers.}
% The objective of generative judgment is to make more accurate judgments between pairs of answers. 
Existing open-source generative judges are generally trained using supervised fine-tuning (SFT)~\citep{kim2024prometheus,zhang2024generative,li2023generative} to imitate correct judgments.
However, we empirically find that only SFT is insufficient~(see Setion~\ref{sec:exp}). 
The intuition is that LLMs should identify the more important aspects of a judgment, rather than patterns that may appear in both positive and negative judgments~\citep{park2024offsetbias}.
For example, both ``answer 1 has logical errors, so the better answer is 2'' and ``answer 2 has logical errors, so the better answer is 1'' could be positive judgments for different prompt inputs, even though they are opposite in meanings.
When LLM is trained with SFT loss, it may primarily imitate the common pattern that appears in both answers rather than developing the ability to make judgments based on the prompt.
Similar studies~\footnote{https://github.com/LLaMafia/SFT\_function\_learning} indicate that the likelihood of generating negative output might even surpass that of positive output during SFT training.

% \passage{Is it beneficial to generate rationales as well as make preference judgments?}
\passage{Rationales bring robustness against bias.} 
\label{sec:rationale}
The proposed Con-J can generate rationales in addition to the binary preference prediction.
% By understanding not only the model's judgments but also its supporting rationales, we can potentially facilitate human-in-the-loop supervision during the LLM alignment step.
We suggest that training the model to generate rationales can impart a regularization effect and help avoid potential biases in the datasets.
Here we provide a theoretical motivation for this effect.
We decompose a judgment $j$ into $j_r$ and $j_y$, representing the rationale and the binary preference prediction, respectively.
Adding the rationales as training targets can be formalized by introducing an intermediate variable $j_r$ influencing the conditional probability $P_{\theta}(j_y \mid p)$:
\begin{equation}
    P_{\theta}(j_y \mid p) = \sum_{j} P_{\theta}(y \mid j_r, p) \, P_{\theta}(j_r \mid p)
\end{equation}
By including rationales, the bias in preference data is distributed between $j_y$ and $j_r$, reducing its direct impact on $j_y$. 
We can formalize the loss function as:
\begin{equation}
\ell(\theta) = - \sum_{(p, j) \in D} \log P_{\theta}(j_y \mid p) - \sum_{(p, j) \in D} \log P_{\theta}(j_r \mid p)
\end{equation}
The loss $P_{\theta}(j_r \mid p)$ encourages the model to find representations that are also effective for predicting $j_r$, as it is less influenced by the bias compared $P_{\theta}(j_y \mid p)$, which solely depends on the dataset's preference annotation.

\passage{Generative judge resists bias with a better prior.}
SM modifies the LLM's architecture with a classification head and uses a discriminative training target. 
On the contrary, Con-J uses an architecture consistent with the pertaining process and generative training objectives~(\cite{zhang2024generative}).
We refer to \citet{erhan2009difficulty} and assume the parameter inherited from the pre-trained LLM as adding an infinite penalty:
\begin{equation}
    \ell(\theta) = \ell_{data}(\theta)  + \frac{\lambda}{2} ||\theta - \theta_0||^2
\end{equation}
where $\lambda$ is the regularization strength.
We make an ideal hypothesis that there exists an optimal $\theta^{*}$ which fully encodes human values and consistently makes true judgments.
Such an $\theta^{*}$ must have sufficient world knowledge beyond what preference datasets can provide during LLM post-training. 
Hence we assume that the parameters obtained during the pre-training phase are closer to $\theta^{*}$ than a random distribution.
This analysis suggests Con-J gets a smaller penalty term for optimization towards $\theta^{*}$. 
In contrast, SM adopts a different training objective and introduces a randomly initialized head, making the regularization effect less significant.
Hence, SM can be good at encoding the knowledge reflected in the preference dataset but is also more sensitive to its bias than Con-J.

\section{Experiments}
\label{sec:exp}
\subsection{Experimental Setup}
\textbf{Datasets.} 
We train scalar models and Con-J on three datasets within different vertical domains: Creation, and Math, then evaluate their performance in terms of the accuracy of preference predictions.
The datasets are self-built commercial datasets consisting of approximately 120,000, 50,000, and 50,000 preference samples for Creation, Math, and Code, respectively.
The Creative dataset involves tasks on text creation such as writing poetry or crafting headlines, whereas Math and Code datasets concentrate on math problem-solving and code writing, respectively.
The datasets cover diverse sources, ranging from data generated by a commercial ChatBot, data generated by ChatGPT, and data from open-source datasets like HH-rlhf~\cite{bai2022training} and Infinity-Instruct\footnote{https://huggingface.co/datasets/BAAI/Infinity-Instruct}.
The preference annotation for these datasets is gathered from human annotators, with each annotation being performed by one annotator and subsequently verified by another.
In addition to the self-built datasets, we train Con-J on a publicly available dataset
Skywork-Reward-Preference-80K-v0.1\footnote{https://huggingface.co/datasets/Skywork/Skywork-Reward-Preference-80K-v0.1} and test its performance on public benchmarks including Infinity-Preference\footnote{https://huggingface.co/datasets/BAAI/Infinity-Preference}, UltraFeedback~\cite{cui2023ultrafeedback}, PKU-SafeRLHF~\cite{ji2024pku}, and Reward-Bench~\cite{lambert2024rewardbench}.
We ensure that no identical prompts appear in both the training and test sets by filtering them out of the training set.
This version of Con-J is publicly released in \textit{anonymized link}.

\textbf{Model.}
We select Qwen2-7B-Instruct~\cite{yang2024qwen2} as the base model to train both the scalar model~(named \textit{\textbf{SM}}) and the proposed generative judge~(named \textit{\textbf{Con-J}}). 
SM includes both pairwise and pointwise variants.
Additionally, we included the original pre-trained Qwen2-7B-Instruct as an untrained variant of Con-J.
In addition, we compare Con-J with a range of generative judges, including GPT-4o\footnote{https://openai.com/index/hello-gpt-4o/} and two generative models~(Auto-J~\cite{li2023generative} and Prometheus 2~\cite{kim2024prometheus}) trained by SFT, Llama series~(Llama3.1-8B, and Llama3.1-70B), and Qwen series~(Qwen2-7B, Qwen2.5-72B).

\textbf{Hyper parameters.}
We train SM and Con-J from with the DeepSpeed library~\cite{rasley2020deepspeed}, Zero Redundancy Optimizer~(ZeRO) Stage 3~\cite{rajbhandari2020zero}, gradient-checkpointing~\cite{chen2016training} and FlashAttention~\cite{dao2022flashattention}.
We use bfloat16 (BF16) and tfloat32 (TF32) mix computation precision.
We set a peak learning rate of $9e-6$ with 3\% warmup steps and cosine scheduler, and a maximum sequence length of 4,096.
The batch sizes of SM and Con-J are set to 128 and 24, respectively.
For Con-J, we linearly combine the SFT loss and the DPO loss with $\alpha = 1e^{-6}$.

\textbf{Sampling and inference strategy for Con-J.} 
We use VLLM~\cite{kwon2023efficient} for the inference for Con-J. 
During repeated sampling and hint-driven sampling, we employ greedy sampling with top-p set at 0.9 and top-k at 20, with a maximum output length set as 512, a temperature of 1.0, and a repetition penalty of 1.2.
During the evaluation, we set top-p at 1.0 and a temperature of 0.0.

\subsection{Main Results}
% \begin{table}[t]
%     \centering
%     \begin{minipage}{0.45\textwidth}
%         \caption{Judgment accuracy of GPT-4o, SM, and Con-J.\label{tab:main1} $*$ indicates the performance difference between Con-J is significant at $p<0.05$ using a pair-wise t-test.}
%         \centering
%         \begin{tabular}{l!{\color{lightgray}\vline}lll}
%         \toprule
%         Model                  & Creation & Math & Code \\ \midrule
%         GPT-4o    & $55.6^{*}$   & $74.8^{*}$ & $68.1^{*}$ \\
%         SM~(point-wise) & $69.4^{*}$          & 84.8 & 69.4 \\ SM~(pair-wise)  & $69.2^{*}$ & 84.6 & 69.6 \\
%         Con-J  & \textbf{72.4} & \textbf{85.0} & \textbf{70.1} \\
%         \bottomrule
%         \end{tabular}
%     \end{minipage}%
%     \hspace{0.05\textwidth} % 控制表格之间的水平间距
%     \begin{minipage}{0.45\textwidth}
%         \caption{Judgment accuracy of Con-J and its variants. \label{tab:main2} $*$ indicates the performance difference between Con-J is significant at $p<0.05$ using a pair-wise t-test.}
%         \centering
%         \begin{tabular}{l!{\color{lightgray}\vline}lll}
%         \toprule
%         Model & Creation & Math & Code \\ \midrule
%         Con-J untrained & $53.6^{*}$   & $63.4^{*}$ & $61.7^{*}$ \\
%         Con-J w/o Hint  & $61.3^{*}$   & $77.4^{*}$ & 68.2 \\
%         Con-J w/o DPO   & $54.6^{*}$  & $64.2^{*}$ & $63.5^{*}$ \\
%         Con-J   & \textbf{72.4}  & \textbf{85.0}   & \textbf{70.1} \\
%         \bottomrule
%         \end{tabular}
%     \end{minipage}
% \end{table}

\begin{wraptable}{r}{0.45\textwidth} % 'r' for right placement, 0.5\textwidth for width
  \centering
  \caption{Judgment accuracy of GPT-4o, SM, and Con-J.\label{tab:main1} $*$ indicates the performance difference between Con-J is significant at $p<0.05$ using a pair-wise t-test.}
        \centering
        \begin{tabular}{l!{\color{lightgray}\vline}lll}
        \toprule
        Model                  & Creation & Math & Code \\ \midrule
        GPT-4o    & $55.6^{*}$   & $74.8^{*}$ & $68.1^{*}$ \\
        SM~(point-wise) & $69.4^{*}$          & 84.8 & 69.4 \\ SM~(pair-wise)  & $69.2^{*}$ & 84.6 & 69.6 \\
        Con-J  & \textbf{72.4} & \textbf{85.0} & \textbf{70.1} \\
        \bottomrule
        \end{tabular}
\end{wraptable}

Table~\ref{tab:main1} presents the results of SM and Con-J on the self-built commercial datasets across three vertical domains.
It can be observed that
(i)~there is no significant difference between the pointwise and pairwise variants of SM. 
Although existing research suggests that concatenating the list of responses improves performance for scoring the responses~\cite{jiang2023llm}, we do not observe this effect on our datasets. 
(ii)~Both Con-J and SM outperform off-the-shelf GPT-4o, indicating that small models trained on domain-specific data can effectively reflect domain-related preferences.
(iii)~On the same preference datasets, Con-J consistently outperforms SM across all tasks with a significant gap on the Text Creation task. 
This indicates that Con-J is more effective at acquiring accurate judgment abilities than SM.
% Future research will analyze the differences in capabilities in the Text Creation task and investigate more task scenarios.

We carry out an ablation study to investigate the variants of Con-J. 
Given that current methodologies often employ Supervised Fine-Tuning (SFT) to train generative judges \cite{li2023generative,kim2024prometheus}, we developed an SFT variant of Con-J, trained exclusively on positive judgments using SFT loss. 
As illustrated in Table~\ref{tab:main2}, Con-J trained with our proposed framework outperforms its variant without DPO loss across all datasets. 
This observation demonstrates the effectiveness of training from contrastive judgments.
In addition, Con-J outperforms its variant without hint-driven sampling, which relies solely on repeated sampling and may be infeasible to construct contrastive judgment pairs for some prompts.
Similar findings have been observed when using self-taught techniques to improve LLMs \cite{zelikman2022star}.
Additional variants of Con-J were also tested, which are detailed in the Appendix.

\begin{table}[t]
\caption{Accuracy of generative judges on the test sets of four benchmarks: Infinity-Preference, UltraFeedback, PKU-SafeRLHF, and Reward-Bench. Results in \textbf{bold} are the best among all models and results with \underline{underline} are the second-best.\label{tab:main_public}} 
\begin{tabular}{l|c!{\color{lightgray}\vline}c!{\color{lightgray}\vline}c!{\color{lightgray}\vline}cccc}
\toprule
& \multirow{2}{*}{\makecell{Infinity-\\Preference}} & \multirow{2}{*}{\makecell{Ultra-\\Feedback}} & \multirow{2}{*}{\makecell{PKU-\\SafeRLHF}} & \multicolumn{4}{c}{\makecell{Reward-Bench}}      \\
&    &    &   & Chat & Chat-H & Safety & Reasoning \\ \midrule
Llama3.1-8B  & 59.0  & 62.9   & 66.4  & 80.7 & 49.8   & 64.0 & 68.1      \\
Llama3.1-70B & 64.0    & 71.4   & 67.6    & \textbf{97.2} & 70.2      & 82.8   & 86.0       \\
Qwen2-7B  & 59.0   & 64.5        & 67.2  &  91.3	  & 44.8	 &  73.6    & 	69.0   \\
Qwen2.5-72B & 70.0 &	66.0 &	58.7 & 86.6	& 61.4 & 74.5 & \textbf{90.7}\\
Auto-J       & 69.0                                   & 63.9                           & 66.9                          & 93.0   & 40.0        & 65.5   & 50.5      \\
Prometheus 2 & 68.0     & 63.3  & 63.0 & 85.5 & 49.1      & 77.1   & 76.5      \\
GPT-4o  & \underline{75.0}  & \underline{72.2}    & \textbf{69.6}  & \underline{95.3} & \underline{74.3}      & \underline{87.6}   & 86.9 \\
Con-J~(ours)  & \textbf{81.0}  & \textbf{73.0} & \underline{68.4} & 91.3 & \textbf{79.6} & \textbf{88.0}   & \underline{87.1} \\
\midrule
\end{tabular}
\end{table}
\begin{wraptable}{r}{0.45\textwidth} % 'r' for right placement, 0.5\textwidth for width
  \centering
  \caption{Judgment accuracy of Con-J and its variants. \label{tab:main2} $*$ indicates the performance difference between Con-J is significant at $p<0.05$ using a pair-wise t-test.}
        \centering
        \begin{tabular}{l!{\color{lightgray}\vline}lll}
        \toprule
        Model & Creation & Math & Code \\ \midrule
        Con-J untrained & $53.6^{*}$   & $63.4^{*}$ & $61.7^{*}$ \\
        Con-J w/o Hint  & $61.3^{*}$   & $77.4^{*}$ & 68.2 \\
        Con-J w/o DPO   & $54.6^{*}$  & $64.2^{*}$ & $63.5^{*}$ \\
        Con-J   & \textbf{72.4}  & \textbf{85.0}   & \textbf{70.1} \\
        \bottomrule
        \end{tabular}
\end{wraptable} 
For a fair comparison with other generative judge models, we trained an open-source version of Con-J using the publicly available dataset Skywork-Reward-Preference-80K-v0.1, and evaluated its performance on public benchmarks, as shown in Table~\ref{tab:main_public}.
Con-J outperforms existing publicly available large language models (LLMs) in the vast majority of benchmarks, including commercial instruction-tuned models such as Llama series, Qwen series, and a series of LLMs trained specially for preference judgments such as Auto-J and Prometheus 2, except in the Chat and Reasoning sub-task of Reward-Bench.
Additionally, Con-J achieves comparable performance with the closed-source model GPT-4o in all benchmarks. 
As of September 18, 2024, Con-J ranks first on the reward-bench leaderboard among all models within 7B parameters, as well as second among open-source generative models.

\subsection{Preference learning yields meaningful and useful rationales.}
% Con-J not only produces a binary judgment but also generates a rationale to support its judgment. 
% Some existing research suggests that rationales or critics generated ahead of preference judgments can facilitate Chain-of-Thought~(CoT) reasoning and help the LLM make better judgments~\cite{lee2024applying,ankner2024critique,ye2024improving,zhang2024generative}.
% Although the generated rationales do not bring about a significant performance increase, we do find that the rationales are also improving during preference learning.
% This indicates that iteratively using the LLM's rationale capacity for preferring learning can also improve its rationale ability.

We select 5 checkpoints trained on different numbers of contrastive judgment pairs, i.e., 2k, 4k, 8k, 16k, and 50k.
Then we prompt GPT-4o with the question, corresponding pair of answers, and the judgment generated by Con-J~(see the prompt in Table~\ref{tab:prompt scoring}).
GPT-4o is instructed to score the judgment ranging from 1 to 5 for the rationale's correctness and from 1 to 3 for the rationale's consistency with its predicted preference. 
Additionally, GPT-4o is tasked as a meta-judge to evaluate whether Con-J makes correct preference predictions. 
In this process, if the judgments of GPT-4o conflict with the dataset's true preference annotations, we exclude these questions for further analysis, as these questions may exceed GPT-4o's capabilities.

\begin{figure}[t]
    \centering
    \begin{subfigure}{0.32\textwidth}
        \centering
        \includegraphics[width=\linewidth]{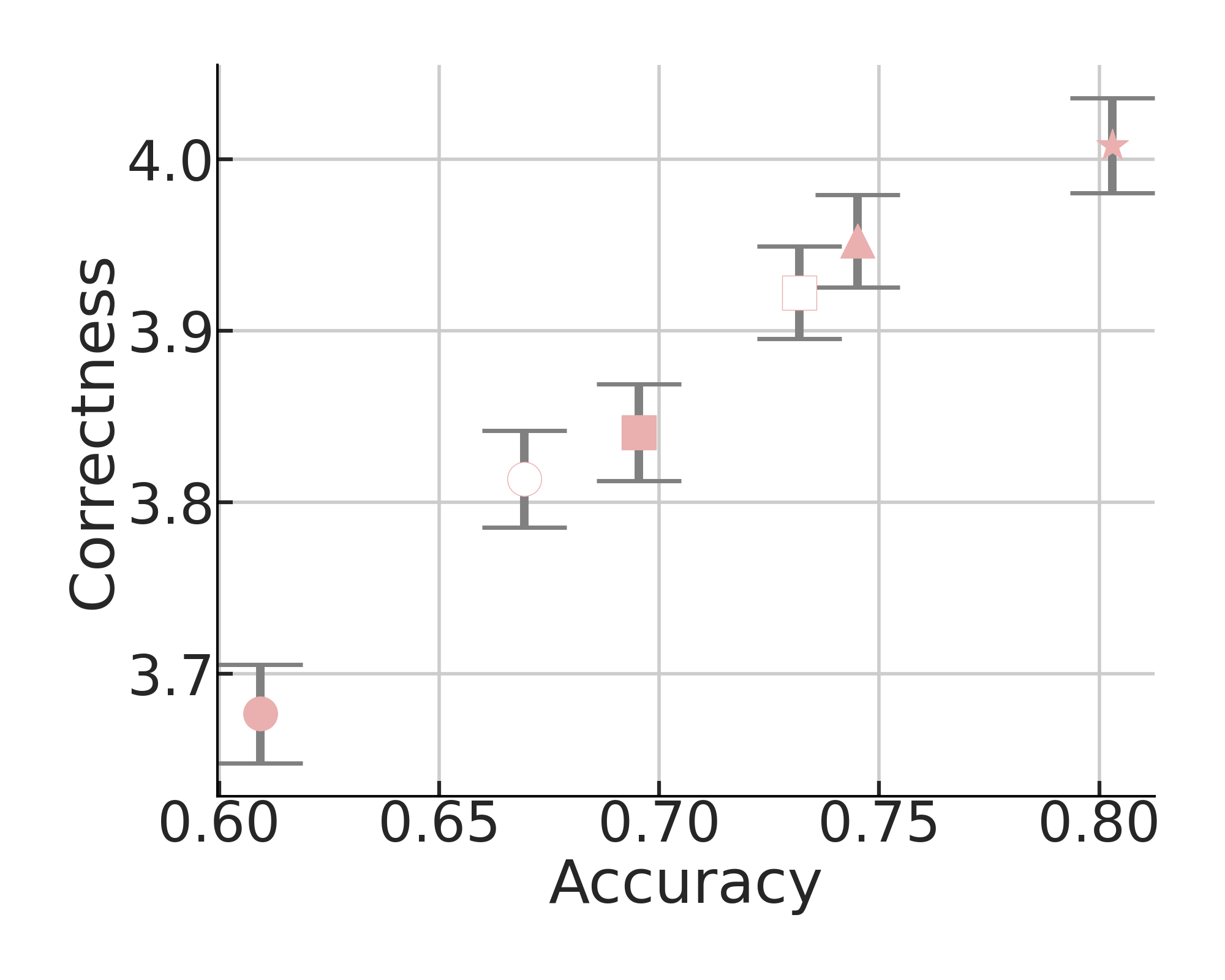}
        \caption{}
    \end{subfigure}%
    \begin{subfigure}{0.32\textwidth}
        \centering
        \includegraphics[width=\linewidth]{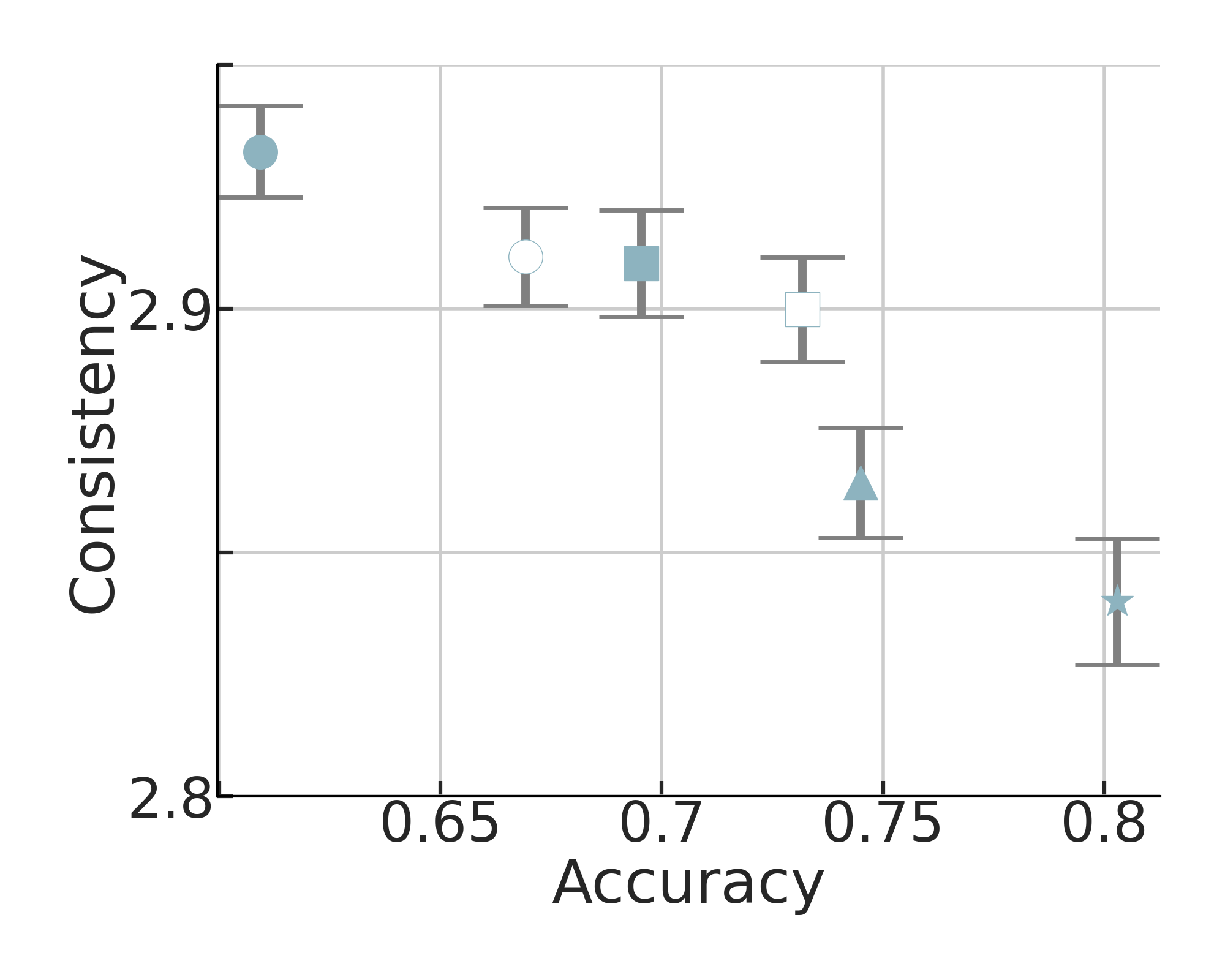}
        \caption{}
    \end{subfigure}%
    \begin{subfigure}{0.32\textwidth}
        \centering
        \includegraphics[width=\linewidth]{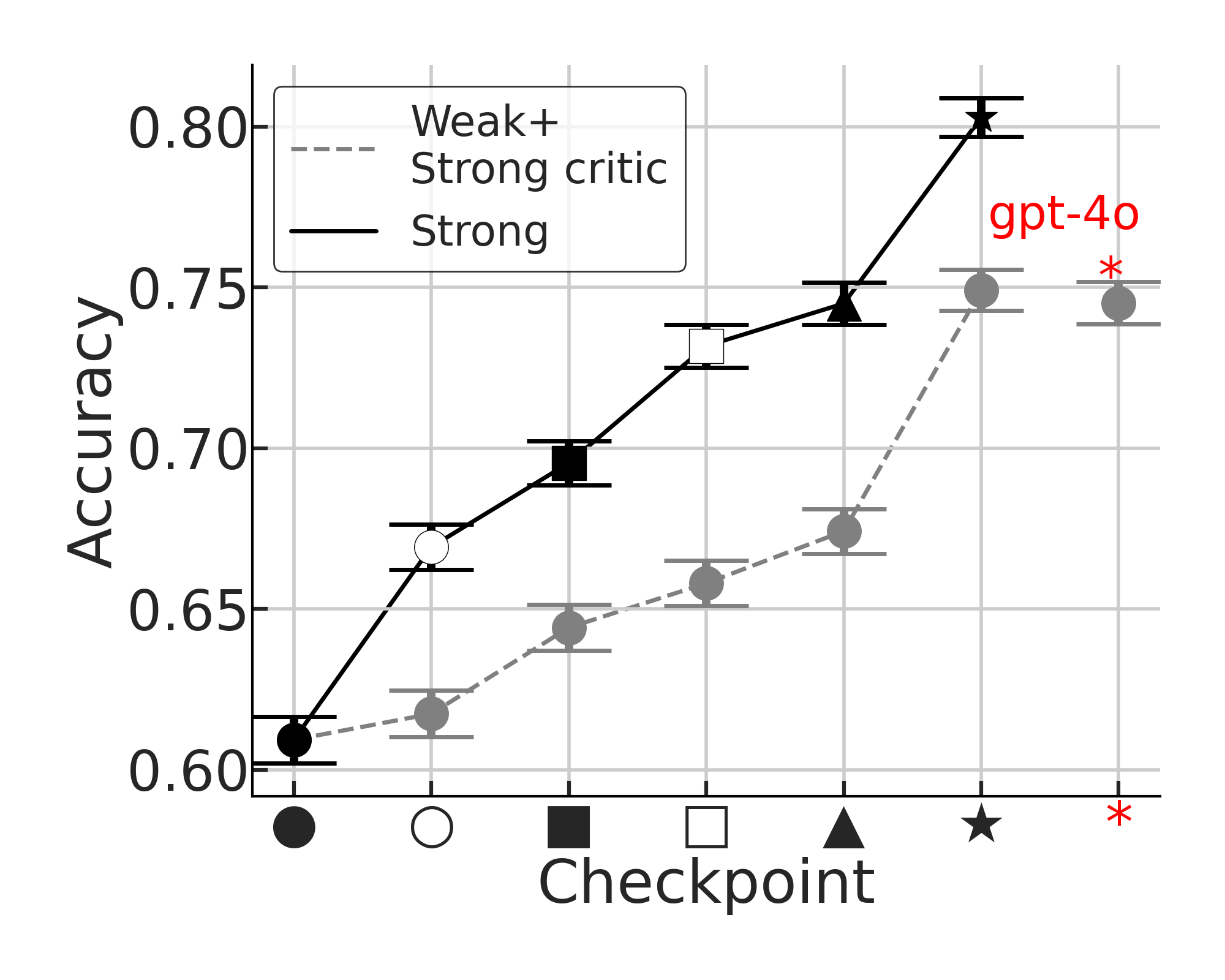}
        \caption{}
    \end{subfigure}
    \caption{ We investigated 5 checkpoints (\ding{108}, \textcircled{ }, \ding{110}, \(\Box\) , \ding{117}, \ding{115}, \ding{116}) trained with different number of contrastive judgment pairs (2k,4k,8k,16k,50k).
    (a-b) We prompt GPT-4o to evaluate the correctness of the rationales and their consistency with the predicted preferences. 
    (c) We tested the use of a strong model's rationale as input provided to a weak model, which can enhance the accuracy of its preference prediction.
    \label{fig:rationale}}
\end{figure}

Experimental results are presented in Figure~\ref{fig:rationale}.
From Figure~\ref{fig:rationale}(a), we observe that the correctness of the rationales improves when Con-J is trained with more data and achieves increased judgment accuracy.
However, we find that the consistency between Con-J's preference prediction and its rationales decreases with the increase in judgment accuracy, as shown in Figure~\ref{fig:rationale}(b).
These observations indicate that Con-J's abilities to make binary preference predictions and generate correct rationales both improve with training. 
However, the increase in inconsistency indicates that these improvements may not be balanced. 
We suspect that because the supervision from preference datasets focuses solely on predicted binary preferences, the enhancement in Con-J's binary prediction ability is more pronounced, leading to inconsistency. 
% This is because  leading to this inconsistency.

We further test whether the improved rationales can be used to help a weak judge make better judgments.
We use the untrained Con-J as the weak judge and prompt it with the rationales generated by the stronger model.
As shown in Figure~\ref{fig:rationale}(c), the weak judge yields more accurate preference prediction with the rationales provided by a stronger model.
Additionally, we find that rationales generated by GPT-4o can also improve the weak judge, with performance comparable to the rationales from the strongest checkpoint of Con-J.
This indicates that Con-J not only surpasses GPT-4o in preference prediction performance but also generates rationales with similar effectiveness.

% Con-J tends to focus more on learning the accuracy of preferential judgments, gradually drifting away from ensuring consistency between the final judgment and the rationale.

\subsection{Rationales help Con-J become less susceptible to biases in the dataset.}
\label{sec:bias}
Scalar models are known to be susceptible to biases in datasets, resulting in preference judgments that reflect biases in the pattern of data rather than true human values. 
To investigate the susceptibility of Con-J, we conduct a synthetic experiment that injects artificial bias into the data.
We define the degree of bias in training set $\mathcal{D}_\text{train}$ as $\gamma$.
This indicates that the $\gamma$ proportion of the data contains unexpected biases that don't necessarily reflect human value, while the rest $1-\gamma$ of the data are randomly sampled from the original training set.

\begin{figure}[t]
    \centering
    \begin{subfigure}{0.255\textwidth}
        \centering
        \includegraphics[width=\linewidth]{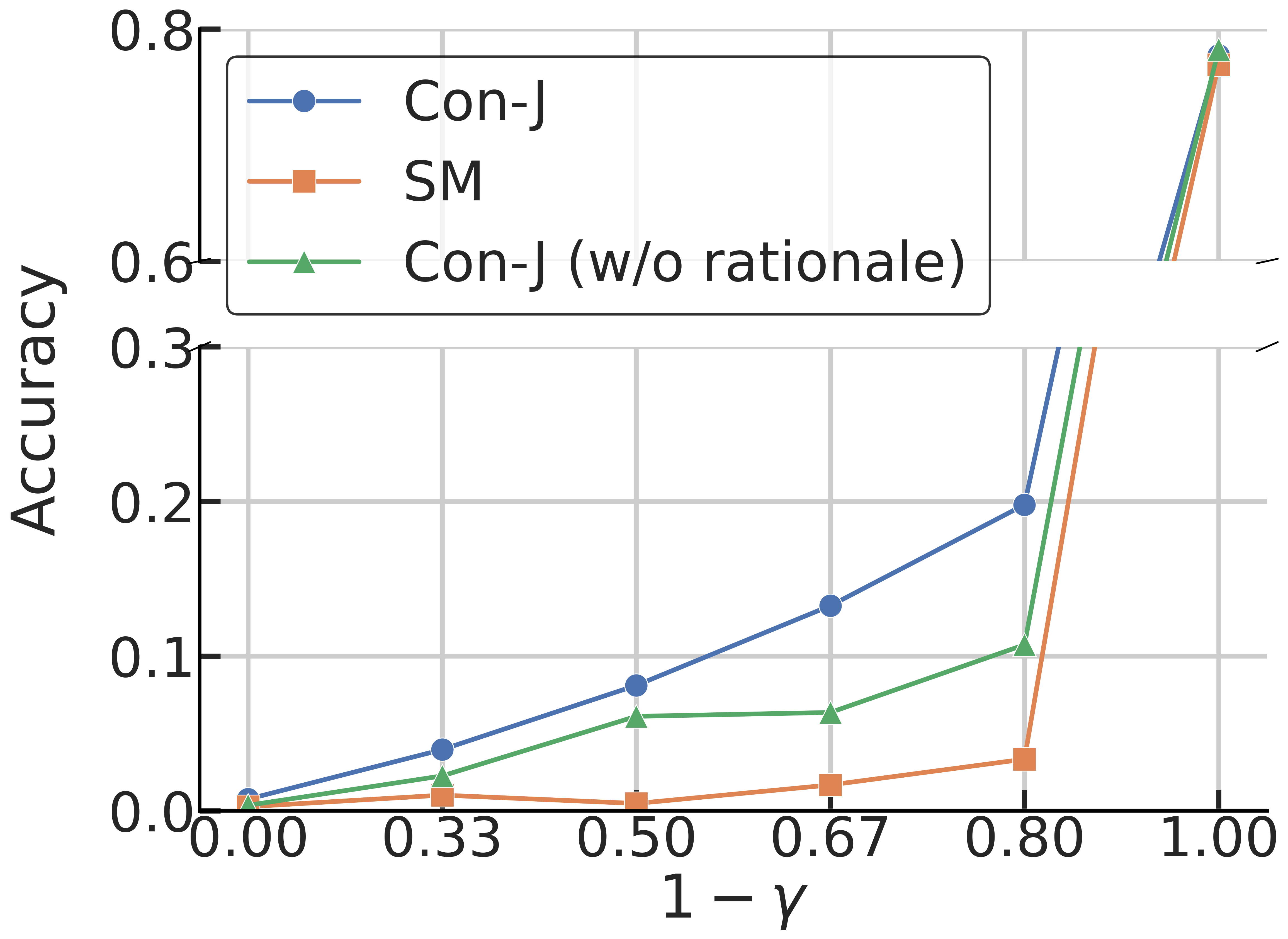}
        \caption{Format (Adversarial)}
    \end{subfigure}%
    \begin{subfigure}{0.25\textwidth}
        \centering
        \includegraphics[width=\linewidth]{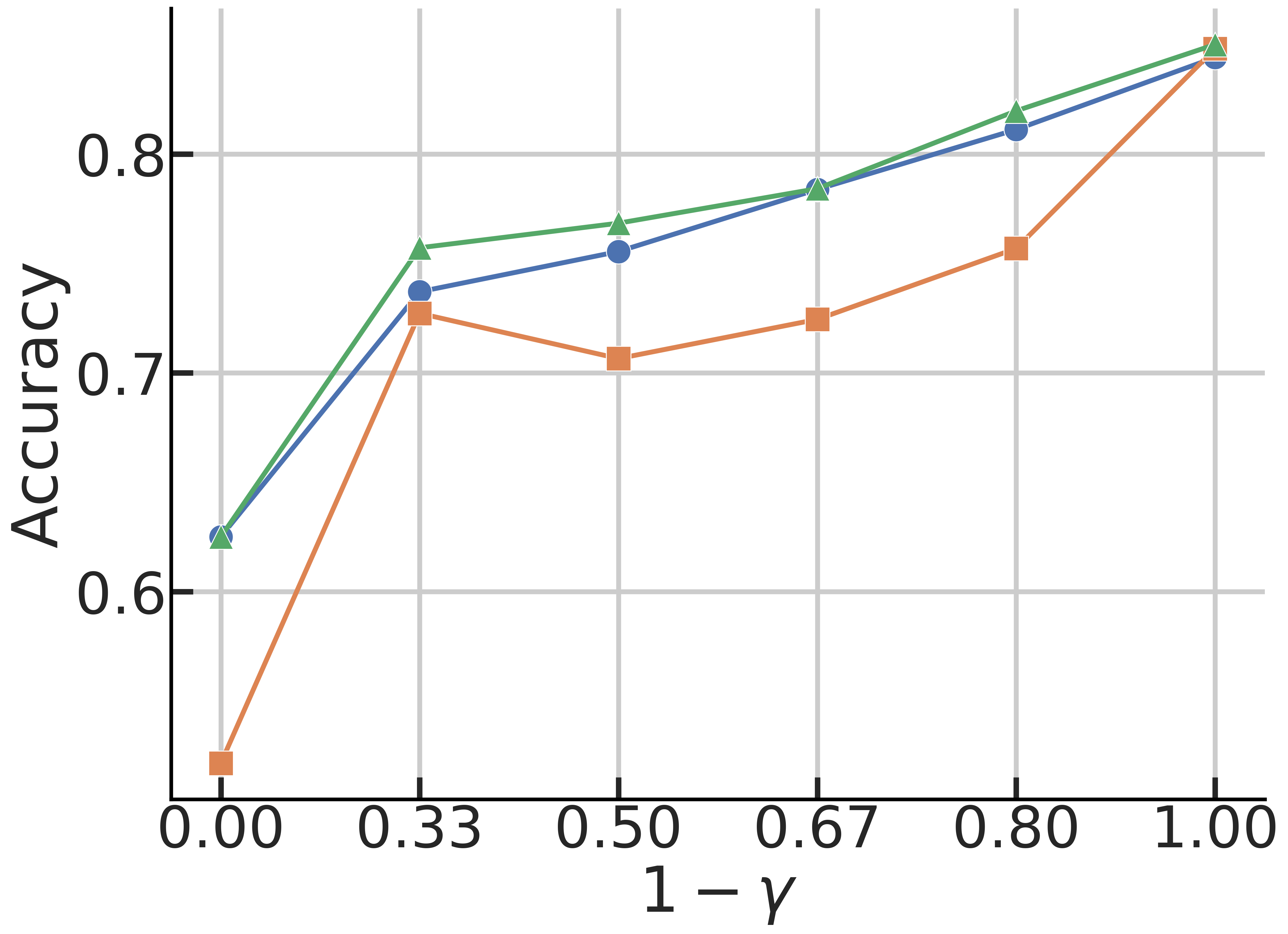}
        \caption{Format (General)}
    \end{subfigure}%
    \begin{subfigure}{0.25\textwidth}
        \centering
        \includegraphics[width=\linewidth]{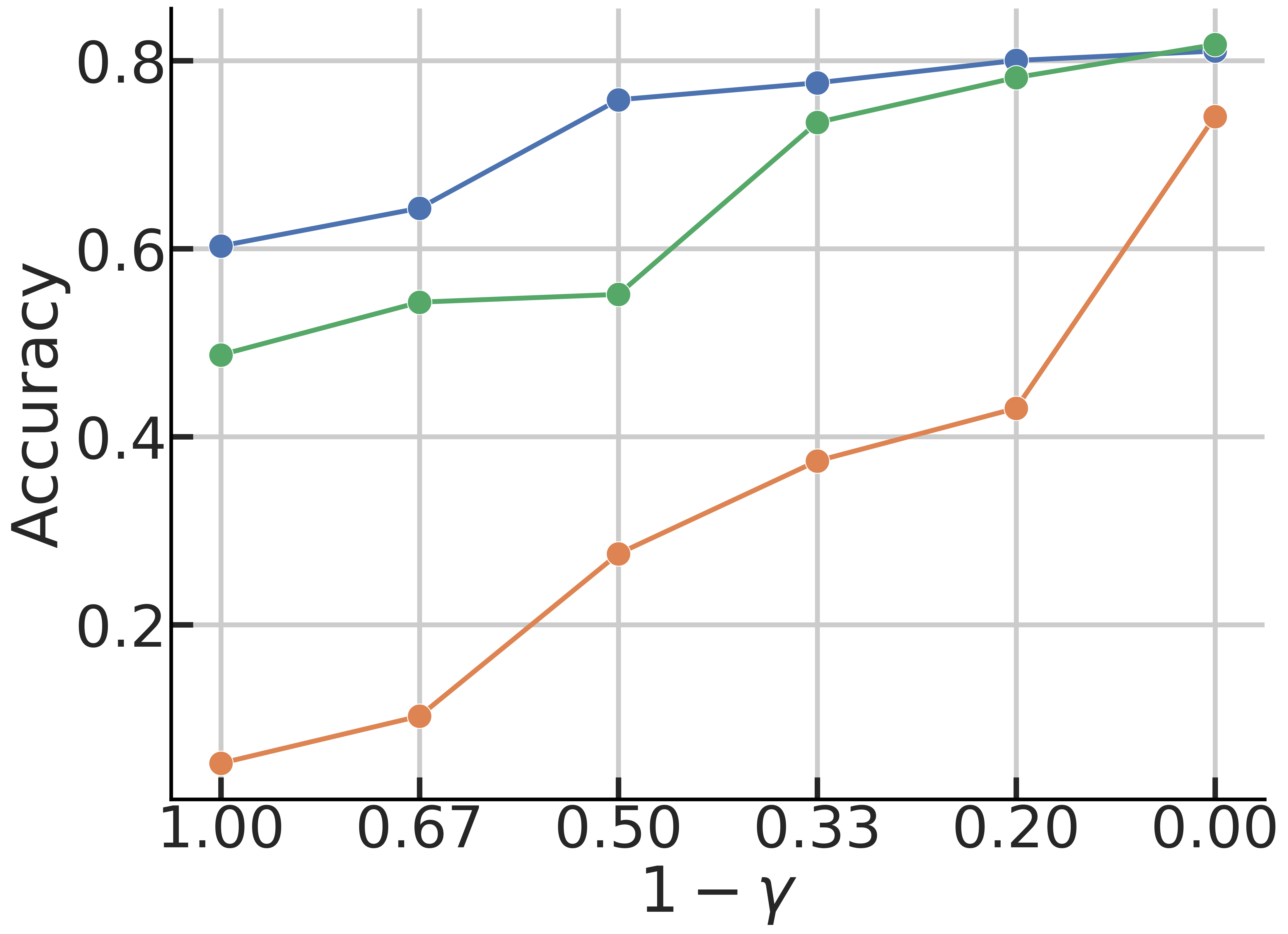}
        \caption{Verbosity (Adversarial)}
    \end{subfigure}%
    \begin{subfigure}{0.25\textwidth}
        \centering
        \includegraphics[width=\linewidth]{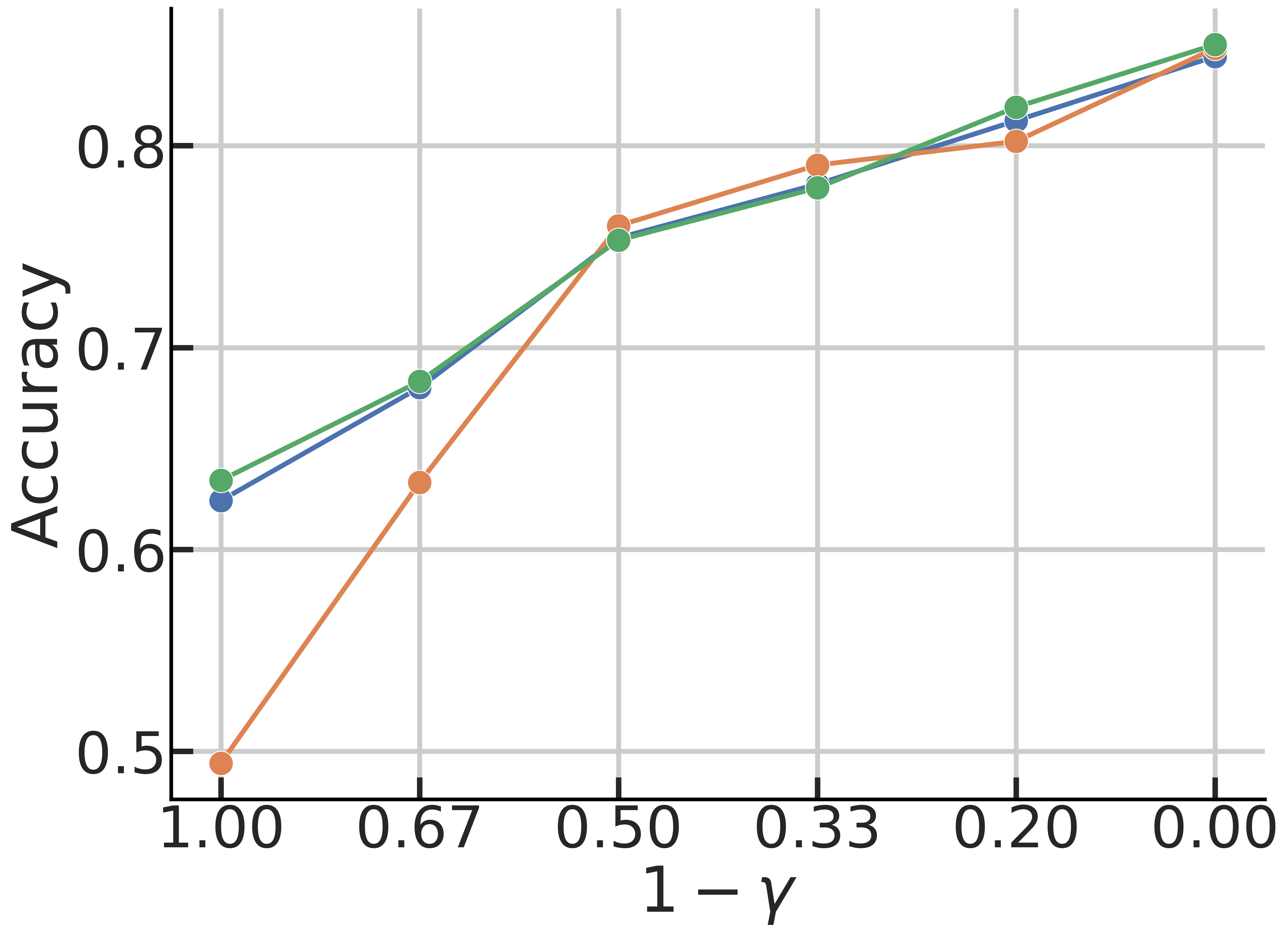}
        \caption{Verbosity (General)}
    \end{subfigure}
    \caption{The performance comparison of Con-J, Scalar model, and Con-J (without rationale), trained with varying degrees of dataset bias~($\gamma$) in format and verbosity, and evaluated on Adversarial and General test sets. \label{fig:format_contrary}}
\end{figure}

We consider two widely studied biases: format bias and verbosity bias~\cite{park2024offsetbias,singhal2023long,shen2023loose}.
For format bias, we use two different prompts to rewrite the pair of answers~(see prompt instructions in Table~\ref{tab:prompt format}). 
Specifically, we rewrite the good answer with a more verbal format and the bad answer with a more formal format with GPT-4o.
% select data from the ground truth data and guarantee that a longer answer is better. 
For verbosity bias, we follow existing practice~\cite{zheng2023judging} by asking GPT-4o to rephrase the good answer without adding any new information and insert the rephrased text at the end of the original answer.

All experiments are conducted on a filtered Math dataset, where the filtering ensures that the solution in the preferred answer is correct, while incorrect in the less preferred answer. 
In this scenario, the format and verbosity of the answers should not be considered in the preference judgments as the correctness of the solution is prioritized.
We investigate the performance of Con-J, SM, and Con-J trained and evaluated without generating the rationales~(see the prompt examples in Table~\ref{tab:prompt no_critic}) under the injections of the addressed bias. 
We trained them on a biased dataset and tested their performance on an original, test set $\mathcal{D}_\text{test}^\text{general}$ without any modification, and a test set constructed by injecting a contrary bias $\mathcal{D}_\text{test}^\text{adversarial}$. 
The contrary bias involves applying the bias for preferred answers from the training set to less preferred answers in the testing phase, and vice versa.
% bias for preferred or less preferred answers in contrast to the one used during the training phase.
The training used the same learning rate as mentioned earlier. 
All models achieved an accuracy of over 99\% on the biased training dataset when $\gamma=1$, indicating the models are thoroughly trained on the data and are strongly influenced by the bias.

We present the experimental results in Figure~\ref{fig:format_contrary}(a-d).
From Figure~\ref{fig:format_contrary}(a-d), we have the following observations:
(i)~Con-J outperforms SM when trained with biased data.
Intuitively, Con-J does not train a value head from scratch; instead, it leverages the text generation capabilities of pre-trained LLMs. 
This approach helps retain global knowledge about the task from pretraining and allows it to extract useful judgment standards even from biased data.
(ii)~When tested on the adversarial datasets, Con-J without rationales significantly underperforms Con-J when $\gamma>0.2$ and $\gamma>0.33$ for the format bias and the verbosity bias, respectively.
The above observations indicate that Con-J becomes more robust at learning from biased data through training with rationales.
(iii)~When tested on general test sets, Con-J without rationales demonstrates comparable performance to Con-J with rationales. 
Some existing research suggests that rationales or critics generated ahead of preference judgments can facilitate Chain-of-Thought (CoT) reasoning and help the LLM make better judgments \cite{lee2024applying, ankner2024critique, ye2024improving, zhang2024generative}. 
A possible explanation for the lack of a similar CoT effect in our data is that the CoT process is often already embedded in the responses, making the CoT procedure for judgment potentially unnecessary.
% This indicates that rationales could potentially 
% Generative methods can effectively learn swimming information from datasets, while requiring result-supported generation can to some extent avoid learning ineffective patterns.

% In Section~\ref{sec:rationale}, we provide a theoretical motivation for this observation by considering the rationale as a regularization effect.
% However, this observation still requires further investigation in the future.
% For example, would there be similar observations in more natural biases, as opposed to artificially introduced biases?
% At the very least, this phenomenon indicates that introducing rationale during the training of generative models is beneficial for counteracting some potential biases.

\subsection{Case Analysis of Con-J}
\begin{table}[t]
\caption{Example of rationales and binary preference generated by Con-J. We prompt GPT-4o to annotate the correctness~(ranging from 1 to 5) of the rationales and consistency between the predicted preference and the rationale~(ranging from 1 to 3). \label{tab:case}}
\scriptsize
\begin{tabular}{c!{\color{lightgray}\vline}p{7.25cm}!{\color{lightgray}\vline}p{3.75cm}!{\color{lightgray}\vline}p{0.7cm}}
\toprule
\multicolumn{2}{c}{\textbf{Prompt input}} & \multicolumn{1}{c}{\textbf{Rationale}} & \multicolumn{1}{c}{\textbf{Preferred}} \\
\midrule
\multirow{1}{*}{\rotatebox[origin=c]{90}{\textbf{Creation~(1)}\phantom{XX}}} & \textbf{Question:} On the theme of comprehensive rural revitalization, propose a paper title. \newline\textbf{Answer 1:} Research on Strategic Pathways for Rural Revitalization from a Multi-Collaborative Perspective. \newline\textbf{Answer 2:} Comprehensive Rural Revitalization: Deep Integration of   Strategies and Practices. & The second answer provides a broader applicability and depth of description, with a smoother and more natural context than the first answer. \newline\textcolor{blue}{\textbf{\textit{{{Correctness: 5, Consistency: 3}}}}} & \vspace{0.15cm} \centering Answer 2 \newline (\textcolor{green}{correct})  \tabularnewline \midrule
\multirow{1}{*}{\rotatebox[origin=c]{90}{\textbf{Math~(1)}\phantom{XXX}}} & \textbf{Question:} Today is August 12, 2021. What was the date 30 days ago?	
\newline\textbf{Answer 1:} According to the given information, today is August 12, 2021, so 30 days ago was July 13, 2021. Therefore, 30 days ago was July 13, 2021.	
\newline\textbf{Answer 2:} July 12, 2021. & The second answer does not provide a detailed explanation, whereas the first answer offers a detailed reasoning process and uses correct date subtraction logic. \newline\textcolor{blue}{\textbf{\textit{{{Correctness: 2, Consistency: 3}}}}} & \vspace{0.15cm} \centering Answer 1 (\textcolor{red}{incorrect}) \tabularnewline \midrule
\multirow{1}{*}{\rotatebox[origin=c]{90}{\textbf{Math~(2)}\phantom{XXXX}}} & \textbf{Question:} Soldiers built a 2000-meter line. They averaged 350 meters per day for the first 3 days. What is the required daily average for the remaining 2 days? \newline
\textbf{Answer1:} $(2000 - 350 \times 3) \div 2 = 1400 \div 2 = 700$ meters.
Answer: An average of 700 meters needs to be set up each day for the last two days. \newline
\textbf{Answer 2:} $(2000 - 350 \times 3) \div 2 = (2000 - 1050) \div 2 = 950 \div 2 = 475$ meters.
Answer: 475 meters for the last two days. & The first answer is straightforward and accurate, taking into account the total distance that needs to be set up and the known information; the second answer offers a slightly simplified expression and the result is correct. \newline\textcolor{blue}{\textbf{\textit{{{Correctness: 2, Consistency: 1}}}}} & \vspace{0.3cm} \centering Answer 2 \newline (\textcolor{green}{correct})\tabularnewline \midrule
\multirow{1}{*}{\rotatebox[origin=c]{90}{\textbf{Code~(1)}\phantom{XXXXXXX}}} & \textbf{Question:}  Reverse the elements in an array and make each character in the elements lowercase.  planet\_arr = {[}`Earth', `Saturn', `Mars'{]}; \newline \textbf{Answer 1:} def   reverse\_array(arr):; return arr{[}::-1{]};  def lowercase\_array(arr):;          return {[}x.lower() for x in arr{]};    planet\_arr = {[}`Earth', `Saturn', `Mars'{]};  print(reverse\_array(planet\_arr));print(lowercase\_array(planet\_arr));\newline
\textbf{Answer 2:} planet\_arr = {[}`Earth', `Saturn', `Mars'{]};     planet\_arr.reverse();      planet\_arr = {[}p.lower() for p in planet\_arr{]}; print(planet\_arr); This code will first reverse the order of the elements in the `planet\_arr` array, and then lowercase each character in each element. The resulting array will be {[}`mars',`saturn',`earth'{]}. & The first answer provides a clearer step-by-step solution to the problem by achieving array reversal and lowercase conversion through independent functions, which makes the code more modular. The second answer is more concise and accurately meets the question's requirements by first reversing the array and then converting it to lowercase. \newline\textcolor{blue}{\textbf{\textit{{{Correctness: 3, Consistency: 1}}}}} & \vspace{0.55cm} \centering Answer 1 (\textcolor{red}{incorrect})  \tabularnewline \bottomrule
\end{tabular}
\end{table}

We present example judgments generated by Con-J in Table~\ref{tab:case}, covering three domains and illustrating both correct and incorrect preference predictions. 
For that cases of Creation~(1) and Math~(1), there is high consistency between the rationale and the binary preference~(the consistency scores are 3). 
We observe that both the rationale and the preference prediction are correct in Creation~(1) while incorrect Math~(1). 
Conversely, the consistency in the cases of Math~(2) and Code~(1) is relatively low. 
In these cases, we find that the rationale does not reflect the most direct support for preference prediction.
For example, in the third example, the answers are compared based on their format, while the key difference is that one of the answers involves an incorrect calculation and is not involved in the rationales.
This indicates that even if the model can make a correct judgment, it may not necessarily be based on the correct rationale. 
% This opens new avenues for human involvement in understanding preference models and devising data to enhance the model's ability not only to make accurate preference judgments but also to base those judgments on correct reasoning.

\section{related work}
\textbf{LLM alignment.}
The initial approach developed for aligning LLMs with human values was reinforcement learning from human feedback~(RLHF)~\citep{christiano2017deep,liu2020learning}.
This technique involves training a scalar reward model~(RM) and then using reinforcement learning~(RL) to optimize a policy according to the RM.
In recent years, a series of \ac{DAP} works, such as DPO~\citep{rafailov2024direct}, SiLC~\citep{zhao2023slic}, and IPO~\cite{azar2024general}, have gained popularity.
Unlike RLHF, DAP methods directly update the LLM using pairwise preference data, making the alignment simpler, and more stable.
To scale the preference datasets, it is common to train an external machine model from existing preference datasets~(\cite{hou2024chatglm,wu2024meta}). 
This online and scalable construction process enables DPO to be deployed in an iterative setting~(\cite{xiong2024iterative, xu2023some}) or online setting~(\cite{guo2024direct}).
Therefore, how to build an accurate external model for preference judgment is an important problem.

\textbf{LLM-as-a-judge.}
Instead of training a scalar model for preference judgment, employing LLMs as a generative judge has been a promising alternative~\citep {zheng2023judging,ye2023flask}.
Efforts have been made to train language models specialized in evaluations.
For example, \citet{li2023generative,kim2024prometheus} proposes constructing an instruction-tuning dataset by prompting GPT-4 and using supervised fine-tuning (SFT) to train a pre-trained LLM as a generative judge. 
\citet{zhang2024generative} propose training the LLM by minimizing the SFT loss with a single ``Yes'' or ``No'', along with a rationale generated either by prompting Gemini 1.0 Pro or through algorithmic construction. 
Our contribution to the existing research is that Con-J uses self-sampled contrastive judgments under the supervision of preference data, allowing for more efficient data construction while achieving better performance.

\section{Discussions and Conclusions}
We introduced Con-J, a novel approach that trains a generative judge by self-bootstrapped learning from preference data. 
Con-J addresses the limitations of scalar reward models, including lack of interpretability and susceptibility to dataset bias.
Our experiments on commercial datasets across Text Creation, Math, and Code domains, as well as publicly available benchmarks, demonstrate the effectiveness of Con-J.  
Moreover, we show that the correctness of the rationales generated by Con-J improves during learning from preference data. 
This enables Con-J not only to make accurate judgments but also to provide reasonable explanations, potentially facilitating human-in-the-loop supervision of LLM alignment.
Finally, we found that Con-J is less susceptible to biases in datasets compared to its variants without rationales and the scalar models.

As AI systems become more powerful, many suggest that they will reach the point at which human are unable to easily and reliably assess the quality of their outputs~\citep{casper2023open}.
To address this issue, using another AI to supervise itself is a viable solution; however, researchers suggest that these methods may fail without human involvement~\citep{shumailov2024ai}.
This paper contributes to addressing this issue in two ways.
On the one hand, Con-J can be used to supervise LLMs by acting as a judge.
At the same time, Con-J produces an explanation of its output that is legible to humans or another trusted system.
This indicates that we can spot any possible errors made by Con-J.
On the other hand, the training and construction of Con-J rely solely on preference data, which is easier to acquire from human annotators than high-quality instruction tunning data.
Furthermore, in many cases humans often find it difficult to provide verbal reasons for their preference, the training of Con-J could be integrated with human preference annotations, thereby enhancing the transparency of the entire annotation process.

Several limitations of this work guide future directions including:
(i)~We demonstrate that preference learning can enhance the model's ability to generate correct rationales. 
Another unresolved and intriguing question is whether enhancing the quality of rationales during the sampling process could also improve the model's preference prediction abilities.
It is an important problem to enhance the model's ability not only to make accurate preference predictions but also to base those judgments on correct reasoning.
(ii)~We demonstrated that Con-J can more effectively resist bias than SM in an adversarial experiment. 
However, further analysis is needed to understand why Con-J outperforms SM on complex, realistic datasets, and whether this is also related to bias.
(iii)~We suggest that Con-j can potentially facilitate human collaboration through interpretable preference judgments for LLM training.
The design of such a pipeline is another interesting and valuable direction.

\bibliography{iclr2025_conference}
\bibliographystyle{iclr2025_conference}

\appendix
\clearpage
\section{Appendix}
% \subsection{Example prompt}
\begin{table}[h]
\caption{An example of a prompt fed to the LLM to generate preference reward judgments, consisting of a preamble~(introduction and instructions describing the task), a question, and a pair of candidate answers. The preamble is neutral and does not indicate which answer is better. \label{tab:prompt example}}
\begin{center}
\begin{tabular}{|p{2cm}|p{11cm}|}
\toprule
Preamble & As an evaluation expert, given a question and its two possible answers, please choose which answer best aligns with coherence, accuracy, coverage, and overall quality. Output your judgment in JSON format, where "rationale" is your explanation, and "better\_answer" is an integer type of 1 or 2, for example, \{``rationale'': ``your explanation'', ``better\_answer'': 1\}. Below are the question and its candidate answers:\\ \midrule
Question & The volume of a rectangular prism is 96 cubic decimeters, and the base area is 16 square decimeters. What is its height in decimeters? \\ \midrule
Answer 1 & We can use the following formula to calculate the height h of the rectangular prism: V = S \textbackslash{}times h. Substitute the known values into the formula: 96 = 16 \textbackslash{}times h; h = 6. Therefore, the height of the rectangular prism is 6 decimeters.\\ \midrule
Answer 2 & 5. Height = Volume / Base Area = 96 / 16 = 5 decimeters.\\ \midrule
Prompt   & \textit{As an evaluation expert, given a question and its two possible answers, please choose which answer best aligns with coherence, accuracy, coverage, and overall quality. Output your judgment in JSON format, where "rationale" is your explanation, and "better\_answer" is an integer type of 1 or 2, for example, \{``rationale'': ``your explanation'', ``better\_answer'': 1\}. Below are the question and its candidate answers:
\newline \textbf{Question:}} The volume of a rectangular prism is 96 cubic decimeters, and the base area is 16 square decimeters. What is its height in decimeters?
\newline \textbf{\textit{Answer 1:}} We can use the following formula to calculate the height h of the rectangular prism: V = S \textbackslash{}times h. Substitute the known values into the formula: 96 = 16 \textbackslash{}times h; h = 6. Therefore, the height of the rectangular prism is 6 decimeters.
\newline  \textbf{\textit{Answer 2:}} 5. Height = Volume / Base Area = 96 / 16 = 5 decimeters. \\ \bottomrule
\end{tabular}
\end{center}
\end{table}

\clearpage
\begin{table}[t]
\caption{Prompt template with preamble using correct or incorrect hints, where $\alpha, \beta$ are the ID of correct and incorrect answers, respectively, $\{\alpha, \beta\}=\{1,2\}$, \{\{Question\}\}, \{\{Answer 1\}\}, \{\{Answer 2\}\} are the text content of the question, answer 1, and answer 2, respectively. When the LLM does not output a valid JSON format as expected (e.g., {``rationale'': ``your explanation'', ``better\_answer'': $\alpha$}), we use an alternative prompt (rows 3-4) to prompt it again and insert its output as the rationale into the template. \label{tab:prompt hint}}
\begin{center}
\begin{tabular}{|p{2cm}|p{11cm}|}
\toprule
Prompt with preamble\_correct  & \textit{As an evaluation expert, given a question and its two possible answers, please choose which answer best aligns with coherence, accuracy, coverage, and overall quality. Below are the question and its candidate answers:
\newline \textbf{Question:}} \{\{Question\}\}
\newline \textbf{\textit{Answer 1:}} \{\{Answer 1\}\}
\newline  \textbf{\textit{Answer 2:}} \{\{Answer 2\}\}
\newline \textit{Given that answer $\alpha$ is better than answer $\beta$, please provide the rationale. Output your judgment in JSON format, where "rationale" is your explanation, and "better\_answer" is an integer type of $\alpha$, for example, \{``rationale'': ``your explanation'', ``better\_answer'': $\alpha$\}}.
\\ \midrule
Prompt with preamble\_incorrect  & \textit{As an evaluation expert, given a question and its two possible answers, please choose which answer best aligns with coherence, accuracy, coverage, and overall quality. Below are the question and its candidate answers:
\newline \textbf{Question:}} \{\{Question\}\}
\newline \textbf{\textit{Answer 1:}} \{\{Answer 1\}\}\newline  \textbf{\textit{Answer 2:}} \{\{Answer 2\}\}
\newline \textit{Given that answer $\beta$ is better than answer $\alpha$, please provide the rationale. Output your judgment in JSON format, where "rationale" is your explanation, and "better\_answer" is an integer type of $\beta$, for example, \{``rationale'': ``your explanation'', ``better\_answer'': $\beta$\}.}\\ \midrule
Prompt with preamble\_correct~(alternative)  & \textit{As an evaluation expert, given a question and its two possible answers, compare the answers according to their coherence, accuracy, coverage, and overall quality. Below are the question and its candidate answers:
\newline \textbf{Question:}} \{\{Question\}\}
\newline \textbf{\textit{Answer 1:}} \{\{Answer 1\}\}\newline  \textbf{\textit{Answer 2:}} \{\{Answer 2\}\}
\newline \textit{Given that answer $\alpha$ is better than answer $\beta$, please provide the rationale:} \\ \midrule
Prompt with preamble\_incorrect~(alternative)  & \textit{As an evaluation expert, given a question and its two possible answers, compare the answers according to their coherence, accuracy, coverage, and overall quality. Below are the question and its candidate answers:
\newline \textbf{Question:}} \{\{Question\}\}
\newline \textbf{\textit{Answer 1:}} \{\{Answer 1\}\}\newline  \textbf{\textit{Answer 2:}} \{\{Answer 2\}\}
\newline \textit{Given that answer $\beta$ is better than answer $\alpha$, please provide the rationale:} \\
\bottomrule
\end{tabular}
\end{center}
\end{table}
\clearpage
\begin{table}[t]
\caption{Prompt template for forcing the generative LLM outputs only the binary judgment without any rationales. \{\{Question\}\}, \{\{Answer 1\}\}, \{\{Answer 2\}\} are the text content of the question, answer 1, and answer 2, respectively. 
% \textbf{Human} and \textbf{Assistant} indicates the role of input content and we force the generative LLM to output 1 or 2  
\label{tab:prompt no_critic}}
\begin{center}
\begin{tabular}{|p{2cm}|p{11cm}|}
\toprule
Prompt  & 
% \textbf{Human:} 
\textit{As an evaluation expert, given a question and its two possible answers, please choose which answer best aligns with coherence, accuracy, coverage, and overall quality. Output your judgment in JSON format in which "better\_answer" is an integer type of 1 or 2, for example, \{``better\_answer'': 1\}. Do not include any additional explanations. Below are the question and its candidate answers:
\newline \textbf{Question:}} \{\{Question\}\}
\newline \textbf{\textit{Answer 1:}} \{\{Answer 1\}\}
\newline  \textbf{\textit{Answer 2:}} \{\{Answer 2\}\}
% \textbf{Assistant:} {``better\_answer'':
\\ 
\bottomrule
\end{tabular}
\end{center}
\end{table}

\clearpage
\begin{table}[t]
\caption{Prompt template for transforming the answers into different formats. 
% \textbf{Human} and \textbf{Assistant} indicates the role of input content and we force the generative LLM to output 1 or 2  
\label{tab:prompt format}}
\begin{center}
\begin{tabular}{|p{2cm}|p{11cm}|}
\toprule
Prompt for rewriting the answer into a more verbal format & 
% \textbf{Human:} 
\textit{You are someone who works on popularizing mathematical knowledge. 
Please restate the following content in simpler, more accessible language without changing the original meaning, affecting its length, or adding extra information. 
Below is the input:
\{\{Answer\}\}}
% \textbf{Assistant:} {``better\_answer'':
\\ \midrule
Prompt for rewriting the answer into a more formal format & 
% \textbf{Human:} 
\textit{You are a researcher in the field of mathematics. 
Please restate the following content using precise mathematical language without changing the original meaning, affecting its length, or adding extra information. 
Below is the input:
\{\{Answer\}\}}
% \textbf{Assistant:} {``better\_answer'':
\\ \midrule
Prompt for rewriting the answer to be more verbose & 
% \textbf{Human:} 
\textit{Please summarize the input by listing the key points in a numbered format. 
Below is the input:
\{\{Answer\}\}}
% \textbf{Assistant:} {``better\_answer'':
\\ 
\bottomrule
\end{tabular}
\end{center}
\end{table}
\clearpage
\begin{table}[t]
\caption{Prompt template for scoring the rationales and judgments of Con-J.   
\label{tab:prompt scoring}}
\begin{center}
\begin{tabular}{|p{11cm}|}
\toprule
\textit{
\#\#Background
}

\textit{
Given below is a question and two corresponding answers:
}

\textit{
\textbf{\textit{Question:}} \{\{Question\}\}
}

\textit{
\textbf{\textit{Answer 1:}} \{\{Answer 1\}\}
}

\textit{
\textbf{\textit{Answer 2:}} \{\{Answer 2\}\}
}

\textit{
A judge has assessed these two answers and judged which one is better. 
Here are its judgment and the corresponding rationales:
}

\textit{
\{\{Judgment\}\}
}

\textit{
\#\#Workflow and Scoring
}

\textit{
Please analyze whether the judgment is correct and evaluate the rationale by scoring them on (i)~correctness and (ii)~consistency with the binary judgment:
}

\textit{
Correctness:
}

\textit{
1: Completely incorrect, with obvious erroneous arguments.
}

\textit{
2: Mostly incorrect, with some correct arguments.
}

\textit{
3: Partially correct, with almost no errors.
}

\textit{
4: Mostly correct.
}

\textit{
5: Completely correct.
}

\textit{
Consistency:
}

\textit{
1: The emotional tone of the reasons is inconsistent with the final judgment.
}

\textit{
2: The reasons have no clear emotional tone.
}

\textit{
3: The emotional tone of the reasons is consistent with the final judgment.
}

\textit{
\#\#Formatting
}

\textit{
Please return the results in JSON format, for example: {``Judgment Correctness'': ``Correct'', ``Rationale Correctness'': 4,  "Rationale Consistency": 3}, where ``Judgment Correctness'' can be either ``Correct'' or ``Incorrect''.}
\\ 
\bottomrule
\end{tabular}
\end{center}
\end{table}

\clearpage

\begin{algorithm}[t]
\caption{Constructing contrastive judgment pairs for Con-J \label{alg1}}
\begin{algorithmic}[1]
\State \textbf{Input:} $\pi$: a pre-trained LLM; a preference dataset $D = \{(q, a^-, a^+)_i\}_{i=1}^N$.
\State \textbf{Output:} $E$: a set 
of constrastive judgment pairs. 
\State $E=\emptyset$
\For{$(q, a^-, a^+)_i \in D$} 
    \State $p=\text{format}(\text{preamble},(q,a^-,a^+)_i)$ \Comment{Prompt construction with preamble}
    \State Get $M(p)$ with repeated sampling \Comment{Judgment generation with repeated sampling}
    \State $M(p)^+, M(p)^- \gets \text{filter\_correct}(M(p)), \text{filter\_incorrect}(M(p))$ \Comment{Selection with ground truth preference}
    \State $E=E\cup \{(j_p,j_n)|j_p\in M(p)^+, j_n \in M(p)^-\}$ 
    \State $p_p,p_n=\text{format}(\text{preamble\_correct},(x,a^-,a^+)_i), \text{format}(\text{preamble\_incorrect},(x,a^-,a^+)_i)$ \Comment{Using preamble with correct or incorrect hint to construct prompt}
    \State Get ${M(p_p)}$,${M(p_n)}$ with hint-driven sampling
    \State $E=E\cup \{(j_p,j_n)|j_p\in M(p_p), j_n \in M(p_n)\}$ 
    \EndFor
\State \textbf{Return} $E$
\end{algorithmic}
\end{algorithm}

\end{document}